\documentclass[runningheads]{llncs}
\usepackage{graphicx}
\usepackage{subcaption}
\usepackage{comment}
\usepackage{cite}
\usepackage{amsmath}
\usepackage{hyperref}
\usepackage{amssymb}
\usepackage{algorithm}
\usepackage{algpseudocode}
\usepackage{booktabs}
\usepackage{tabularx}
\usepackage{multirow}
\usepackage{marvosym}
\usepackage{acronym}
\usepackage{orcidlink}

\hypersetup{
    colorlinks = true,
    linkcolor = blue,
    urlcolor = blue,
    citecolor = blue
}
\usepackage{enumitem}

\begin{document}

\acrodef{TKS}[TKS]{Text Kernel Stretching}
\acrodef{LEM}[LEM]{Layout Enhanced Module}
\acrodef{IEDP}[IEDP]{Iterative Expansion Distance Post-processor}
\acrodef{SOTA}[SOTA]{state-of-the-art}
\acrodef{CHDAC}[CHDAC]{IACC2022CHDAC}
\acrodef{HDRC}[HDRC]{ICDAR2019HDRC Chinese}
\acrodef{NMS}[NMS]{Non-Maximum Suppression}
\acrodef{LEB}[LEB]{Layout Enhanced Blocks}
\acrodef{FFN}[FFN]{Feed-Forward Network}
\acrodef{TKH}[TKS]{Tripitaka Koreana in Han}
\acrodef{MTH}[MTH]{Multiple Tripitaka in Han}
\acrodef{IoU}[IoU]{Intersection over Union}
\acrodef{FPN}[FPN]{Feature Pyramid Network}
\acrodef{OCR}[OCR]{Optical Character Recognition}
\title{SegHist: A General Segmentation-based Framework for Chinese Historical Document Text Line Detection}
\titlerunning{SegHist}

\author{Xingjian Hu\inst{1}\orcidlink{0009-0000-8258-975X} \and
Baole Wei\inst{1} \and
Liangcai Gao\inst{1}(\Letter) \and
Jun Wang\inst{2}}
\authorrunning{X. Hu et al.}
%
\institute{Wangxuan Institute of Computer Technology, Peking University, Beijing, China
\email{\{huxingjian, gaoliangcai\}@pku.edu.cn}\\
\and
Department of Information Management, Peking University, Beijing, China\\
}

\maketitle              

\begin{abstract}
Text line detection is a key task in historical document analysis facing many challenges of arbitrary-shaped text lines, dense texts, and text lines with high aspect ratios, etc. In this paper, we propose a general framework for historical document text detection (SegHist), enabling existing segmentation-based text detection methods to effectively address the challenges, especially text lines with high aspect ratios. Integrating the SegHist framework with the commonly used method DB++, we develop DB-SegHist. This approach achieves \ac{SOTA} on the \ac{CHDAC}, MTHv2, and competitive results on \ac{HDRC} datasets, with a significant improvement of 1.19\% on the most challenging \ac{CHDAC} dataset which features more text lines with high aspect ratios. Moreover, our method attains \ac{SOTA} on rotated MTHv2 and rotated \ac{HDRC}, demonstrating its rotational robustness. The code is available at \url{https://github.com/LumionHXJ/SegHist}.

\end{abstract}
\setcounter{footnote}{0} 

\section{Introduction}
\begin{figure}[htbp]
    \centering
    \begin{subfigure}{0.24\textwidth}
        \centering
        \includegraphics[width=\textwidth,height=0.67\textwidth]{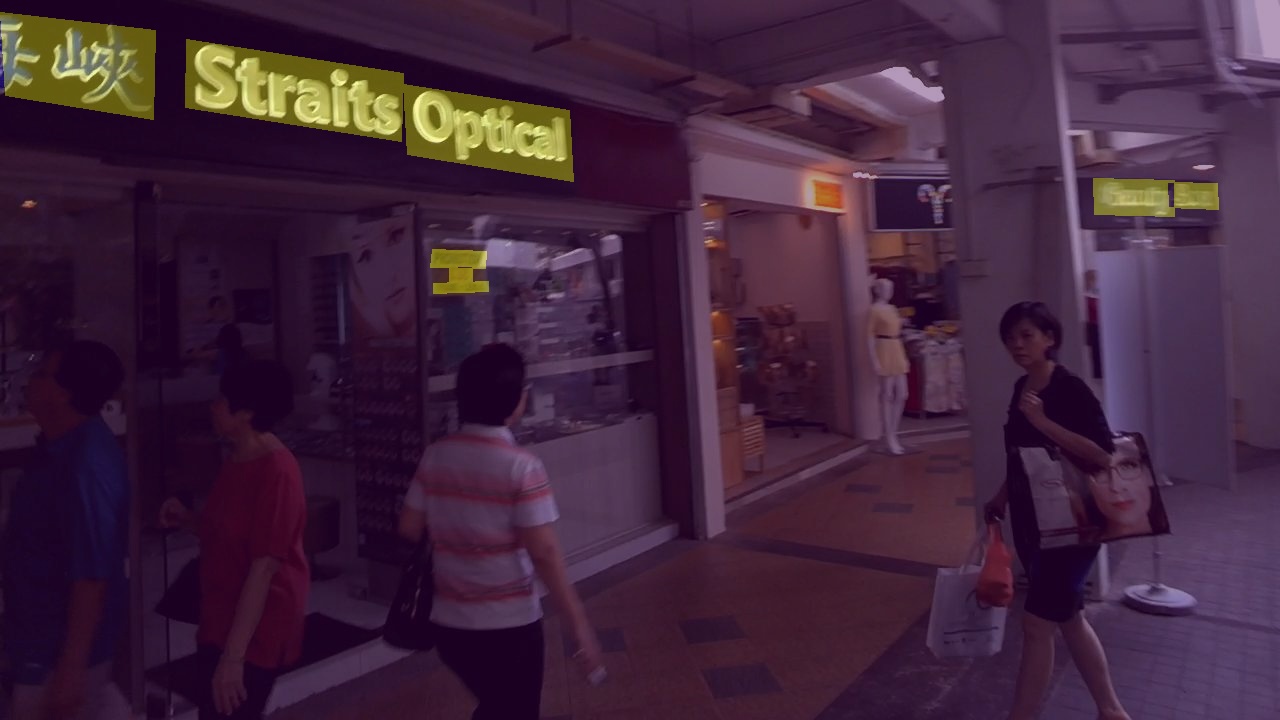}
        \caption{}
        \label{fig:intro-icdar}
    \end{subfigure}
    \hfill
    \begin{subfigure}{0.24\textwidth}
        \centering
        \includegraphics[width=\textwidth,height=0.67\textwidth]{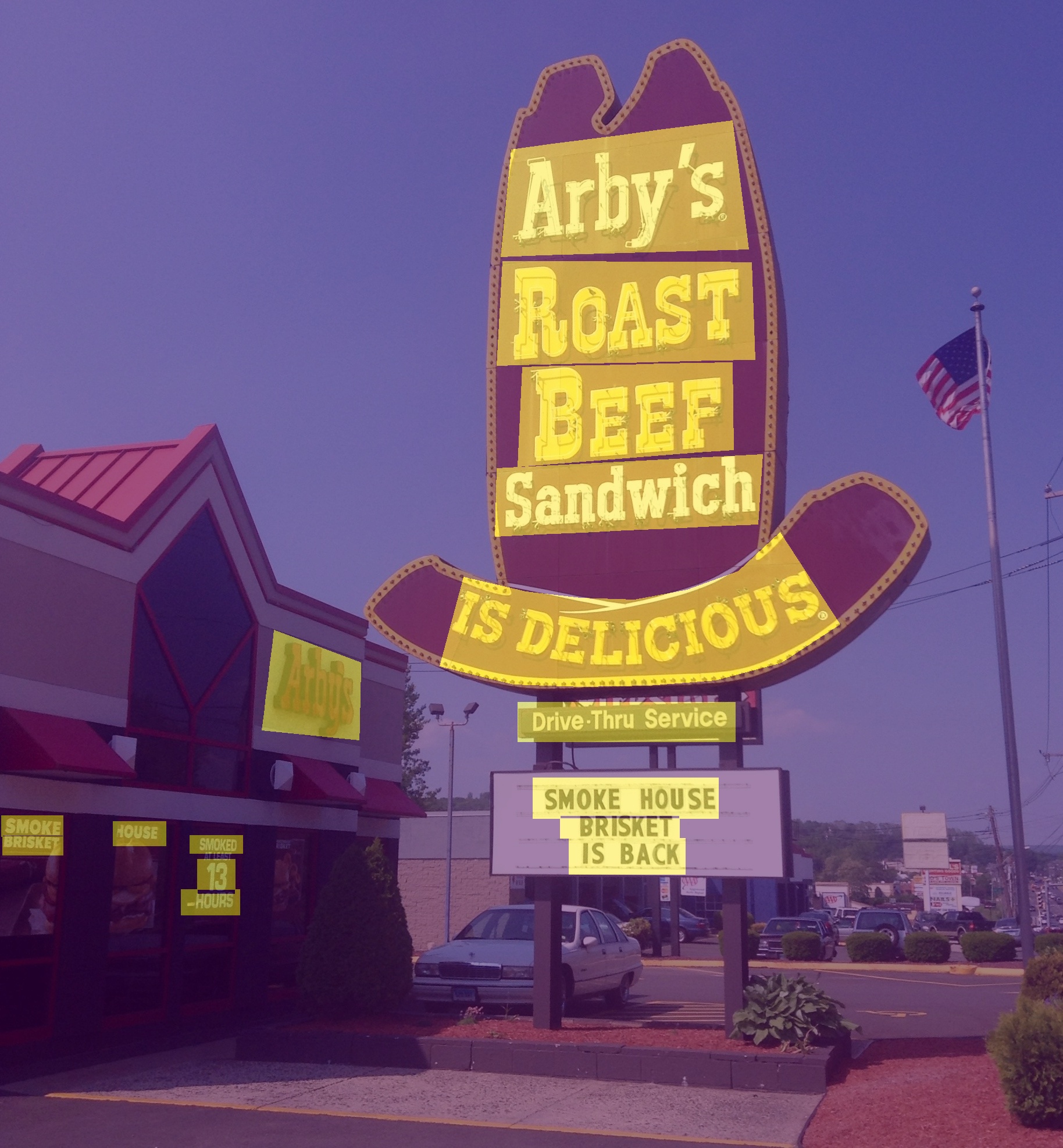}
        \caption{}
        \label{fig:intro-ctw1500}
    \end{subfigure}
    \hfill
    \begin{subfigure}{0.24\textwidth}  
        \centering
        \includegraphics[width=\textwidth,height=0.67\textwidth]{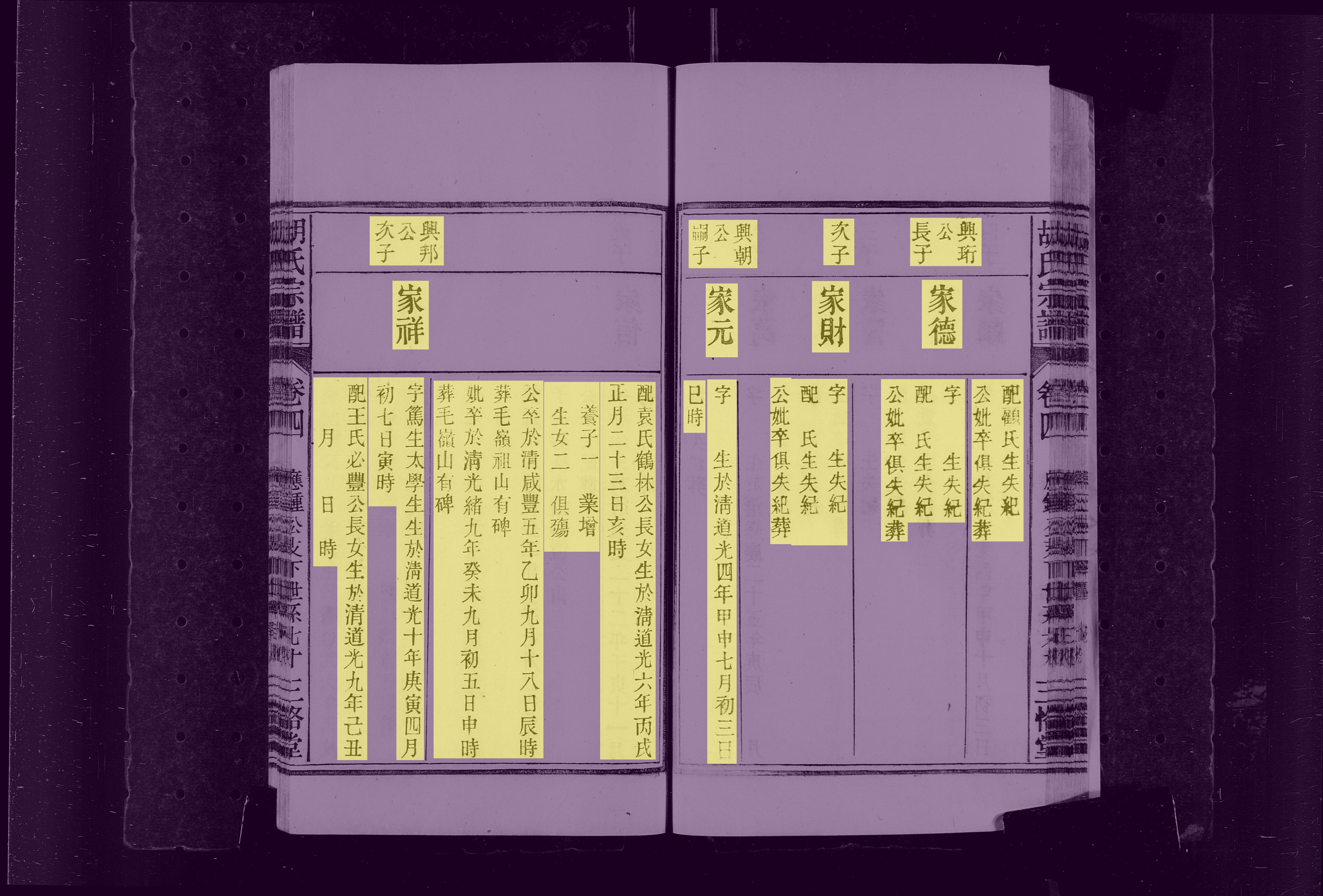}
        \caption{}
        \label{fig:intro-icdar19}
    \end{subfigure}
    \hfill
    \begin{subfigure}{0.24\textwidth}
        \centering
        \includegraphics[width=\textwidth,height=0.67\textwidth]{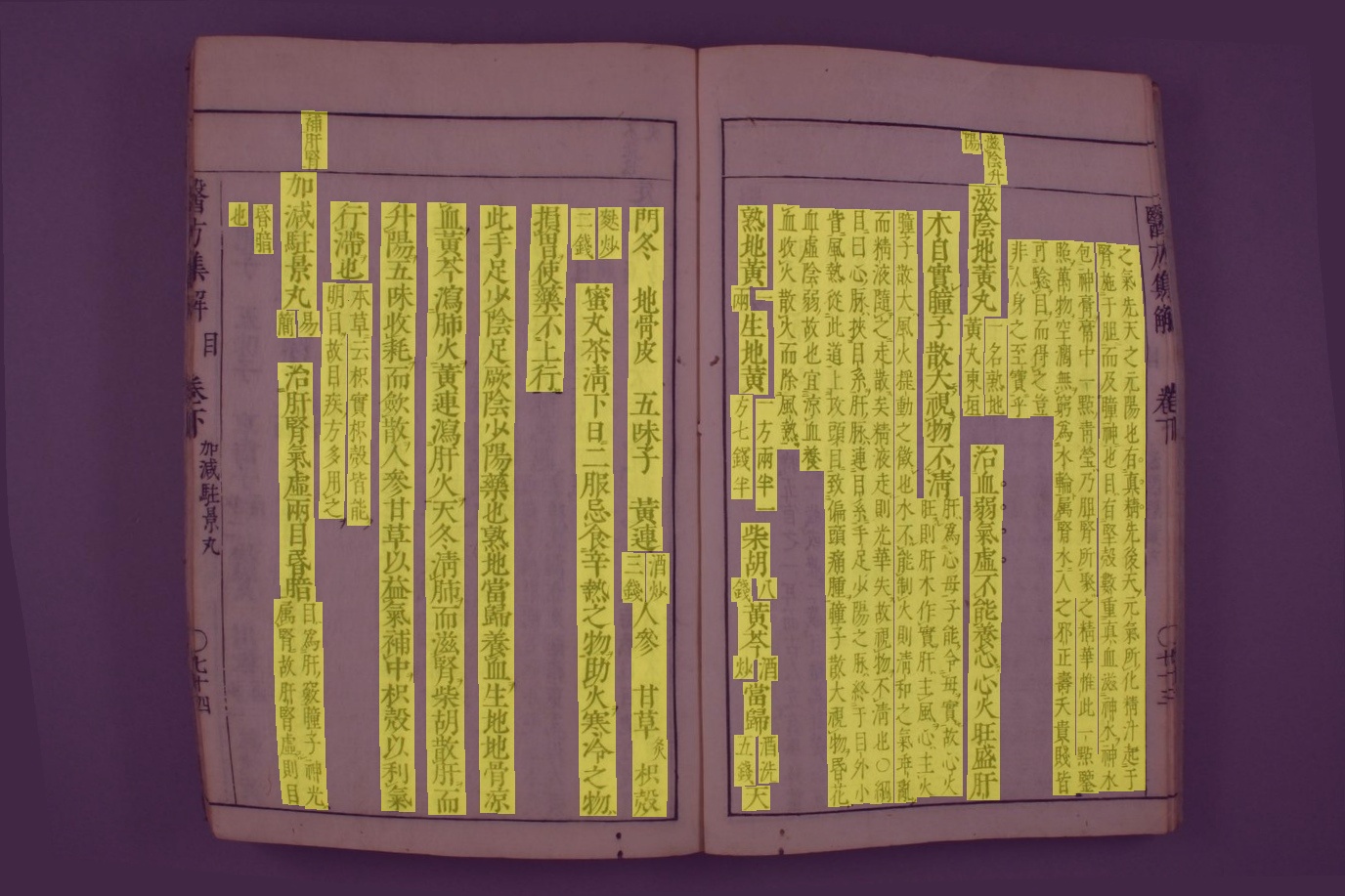}
        \caption{}
        \label{fig:intro-chdac}
    \end{subfigure}
    \caption{Text regions of examples from scene text datasets and historical document datasets. (a) ICDAR2015~\cite{karatzas2015icdar}, (b)SCUT-CTW1500~\cite{yuliang2017detecting}, (c)HDRC~\cite{saini2019icdar}, (d)CHDAC.}
    \label{fig:intro}
\end{figure}

Historical documents analysis~\cite{cheng2022scut,ma2020joint,saini2019icdar,sihang2020precise,xie2019weakly} is pivotal for preserving and disseminating historical documentary materials. Accurately detecting text line positions is essential for downstream tasks like text recognition and layout understanding. As shown in Fig.~\ref{fig:intro}, compared to texts in natural scenes, texts in historical documents are denser, and historical documents contain more text lines with high aspect ratios\footnote{Here, the aspect ratio of a text line refers to the ratio of its longer side to its shorter side. Polygonal text areas, considered as deformed rectangles, have their side lengths calculated through polyline lengths.}. Recent works~\cite{li2023dtdt, jian2023hisdoc} have improved existing text detection methods to handle arbitrary-shaped text lines in historical documents. However, these methods lack effective improvements for the characteristic of text lines and utilize the rich layout structural information neither. 

Segmentation-based methods (\textit{seg-based methods} for short) can represent text lines of arbitrary shapes flexibly, and are highly related to layout segmentation. Unlike regression-based methods (\textit{reg-based methods} for short) that require generating a bounding box for each text line, seg-based methods only need to predict a segmentation map and extract text instances from it, making them suitable for handling numerous text lines in historical documents. Nevertheless, seg-based methods still face difficulties in detecting text in historical documents, particularly in identifying dense texts and text lines with high aspect ratios. 

To address these challenges, we extend existing seg-based methods to the historical document environment. We propose a general framework for historical document text detection tasks, named SegHist. SegHist includes the label generation method \ac{TKS}, which can generate segmentation map targets that are easier to learn and predict, targeting the characteristics of text lines in historical document data; the \ac{LEM} for capturing global information, enhancing the model's ability to access the rich layout structural information in historical documents; and the \ac{IEDP}, a hyperparameter-free post-processing method that optimizes the shortcomings of low accuracy when directly recovering text regions with high aspect ratios.

The contributions of this paper can be summarized as follows:
\begin{enumerate}[topsep=0pt, partopsep=0pt]
\setlength{\itemsep}{0pt}
\setlength{\parskip}{0pt}
	\item In response to text lines with high aspect ratios in historical documents, we propose the general seg-based framework SegHist for historical document text detection.
	\item We integrate SegHist to DBNet++~\cite{liao2022real}, resulting in DB-SegHist, which achieves \ac{SOTA} on the \ac{CHDAC}, MTHv2, and competitive results on \ac{HDRC} datasets, with a significant improvement of 1.19\% on the most challenging \ac{CHDAC} datasets, confirming its effectiveness in complex historical documents feature more text lines with high aspect ratios.
        \item DB-SegHist attains \ac{SOTA} on rotated MTHv2 and rotated \ac{HDRC} datasets, which demonstrating its robustness.
\end{enumerate}

\section{Related Work}
\subsection{Scene Text Detection Methods}
Before the era of deep learning, MSER~\cite{matas2004robust} and SWT~\cite{epshtein2010detecting} were mainstream text detection methods that analyzed images by finding correlations in pixel values. In recent years, deep learning methods have achieved good results in scene text detection tasks, surpassing traditional methods. These methods can generally be divided into reg-based methods and seg-based methods.

\subsubsection{Regression-Based Methods}
Reg-based methods approach the problem of text detection as a special case of object detection, obtaining the representation of text instances by directly predicting the control points of bounding boxes. Textboxes~\cite{liao2017textboxes} applies the SSD~\cite{liu2016ssd} framework to text detection by designing anchor boxes with various aspect ratios and incorporating asymmetric convolutional kernels. EAST~\cite{zhou2017east} predicts the quadrangular bounding boxes with high efficiency, enabling accurate detection of inclined text lines. To address the challenge of curved text, ABCNet~\cite{liu2020abcnet,liu2021abcnet} represents text regions by Bezier curves. Inspired by Mask R-CNN~\cite{he2017mask}, the Mask Textspotter series~\cite{lyu2018mask,liao2020mask} introduces a mask branch to predict curved text regions more accurately. 

However, most of these methods require manual setting of anchors to accommodate the scales and aspect ratios of text lines and a \ac{NMS} operation. Drawing on the DETR~\cite{carion2020end} and Deformable DETR~\cite{zhu2020deformable}, TESTR~\cite{zhang2022text} and DPText-DETR~\cite{ye2023dptext} directly predict a specified number of bounding boxes, thereby avoiding these shortcomings. However, they still face difficulties in detecting images with numerous text lines, a common feature in historical document images.

\subsubsection{Segmentation-Based Methods} 
Seg-based methods view text detection as a special case of semantic segmentation by predicting whether each pixel belongs to a text region, followed by specific post-processing methods. TextSnake~\cite{long2018textsnake} describes text by using a series of discs along the text center line. PSENet~\cite{wang2019shape} utilizes multiple shrinkage levels in prediction and expands text regions during post-processing. PAN~\cite{wang2019efficient} additionally predicts a similarity vector, grouping pixels with higher similarity to a text kernel. DB~\cite{liao2020real} and DB++~\cite{liao2022real} simplify the process by introducing differentiable binarization to create a binary map and recover the text instances by unclipping text kernels directly. With their capability to represent texts of arbitrary shapes and effectiveness in handling numerous text lines, seg-based methods are well-suited for historical document text detection.

\subsection{Historical Document Text Line Detection Methods}
Text detection in historical documents is closely related to scene text detection. Droby et al.~\cite{droby2022text} demonstrated good adaptability in Arabic historical manuscripts containing various diacritical marks using Mask R-CNN~\cite{he2017mask}. SeamFormer~\cite{vadlamudi2023seamformer} uses ViT~\cite{dosovitskiy2020image} to process historical palm leaf manuscripts in two stages to fit text line polygons. Rahal et al.~\cite{rahal2023layout} proposed a lightweight network, L-U-Net, based on FCN~\cite{long2015fully} for historical document analysis. Although these methods have achieved certain results, their direct application to Chinese historical documents, which often differ significantly in text line characteristics and layout structure, is challenging.

In the task of text line detection in Chinese historical documents, Ma et al.~\cite{ma2020joint} enhanced the Faster R-CNN framework~\cite{ren2015faster} by adding a character prediction branch, achieving commendable results on the MTHv2 dataset. HisDoc R-CNN~\cite{jian2023hisdoc} applies Mask R-CNN approach~\cite{he2017mask}, employing iterative methods inspired by Cascade R-CNN~\cite{cai2018cascade} to refine predictions and accurately detect distorted text lines. DTDT~\cite{li2023dtdt} integrates the DETR structure~\cite{carion2020end}, adding a mask representation based on discrete Fourier transform, and thus achieves predictions for arbitrarily shaped texts in historical documents. These methods have achieved commendable results, but they lack improvements in the aspect ratio characteristics of text lines and do not fully utilize the layout structure information in historical documents.

\section{Methodology}
\begin{figure}[htbp]
    \centering
    \includegraphics[width=\textwidth]{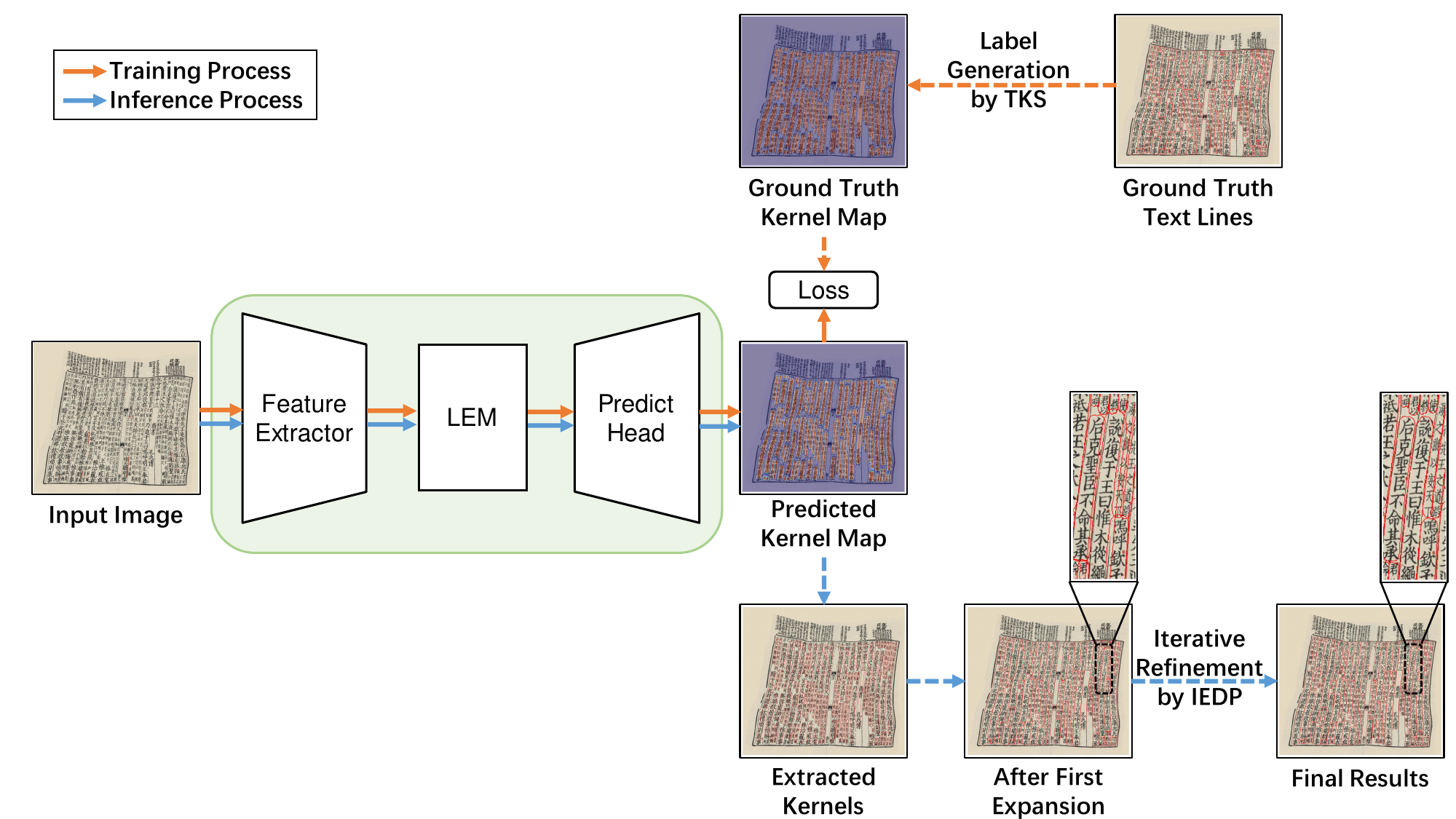}
    \caption{The pipeline of training and inference process of SegHist. The framework is required to integrate into seg-based methods to achieve a commendable performance, which will be described in Section~\ref{sec:integrate}.}
    \label{fig:pipeline}
\end{figure}

\subsection{Overall Pipeline}
To adapt seg-based methods for historical document text detection, we design a general framework, named SegHist. It includes a target generation method (\ac{TKS}), a global information-capturing module (\ac{LEM}), and an iterative post-processing method (\ac{IEDP}). During training, images are passed into a feature extractor that consists of ResNet-50~\cite{he2016deep} and \ac{FPN}~\cite{lin2017feature}, followed by a feature fusion module that upsamples the pyramid features. This process yields features \( F \in \mathbb{R}^{C \times H_0 \times W_0} \), where \( H_0 \) and \( W_0 \) are one quarter of the input image's height and width, respectively. Features \(F\) and learnable layout tokens \(z \) are then fed into the \ac{LEM} and passed to the prediction head to obtain the predicted kernel map. The loss function is computed using this map and the kernel map generated by \ac{TKS}. During inference, text kernels are extracted from the predicted kernel map and are efficiently and accurately recovered through \ac{IEDP}. The pipeline of our framework is shown in Fig.~\ref{fig:pipeline}.

\subsection{Text Kernel Stretching}
\label{sec:tks}

\begin{figure}[htbp]
    \centering
    \begin{subfigure}{0.24\textwidth}
        \includegraphics[width=\textwidth]{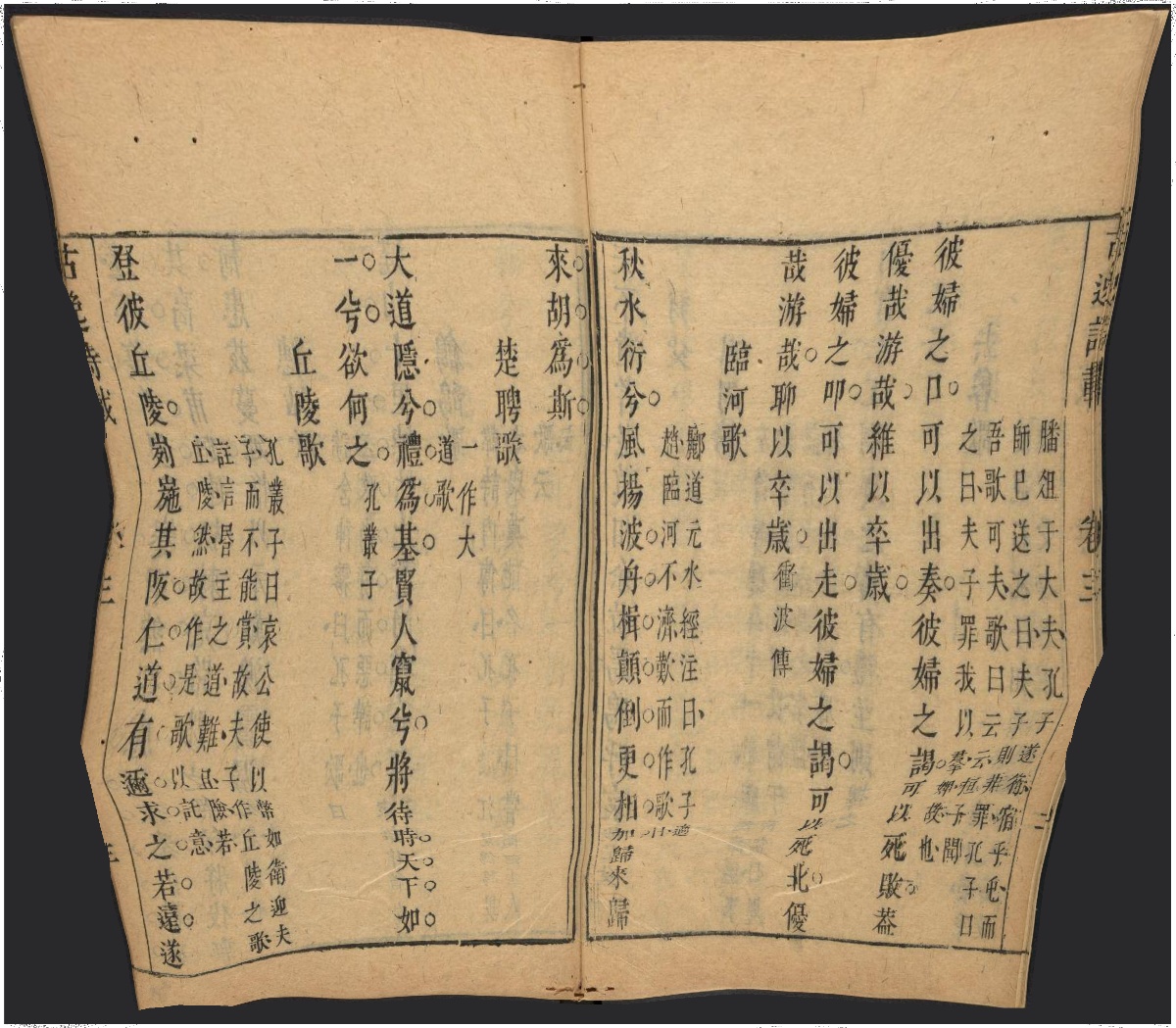}
        \caption{}
        \label{fig:kernel_map_image}
    \end{subfigure}
    \hfill
    \begin{subfigure}{0.24\textwidth}
        \includegraphics[width=\textwidth]{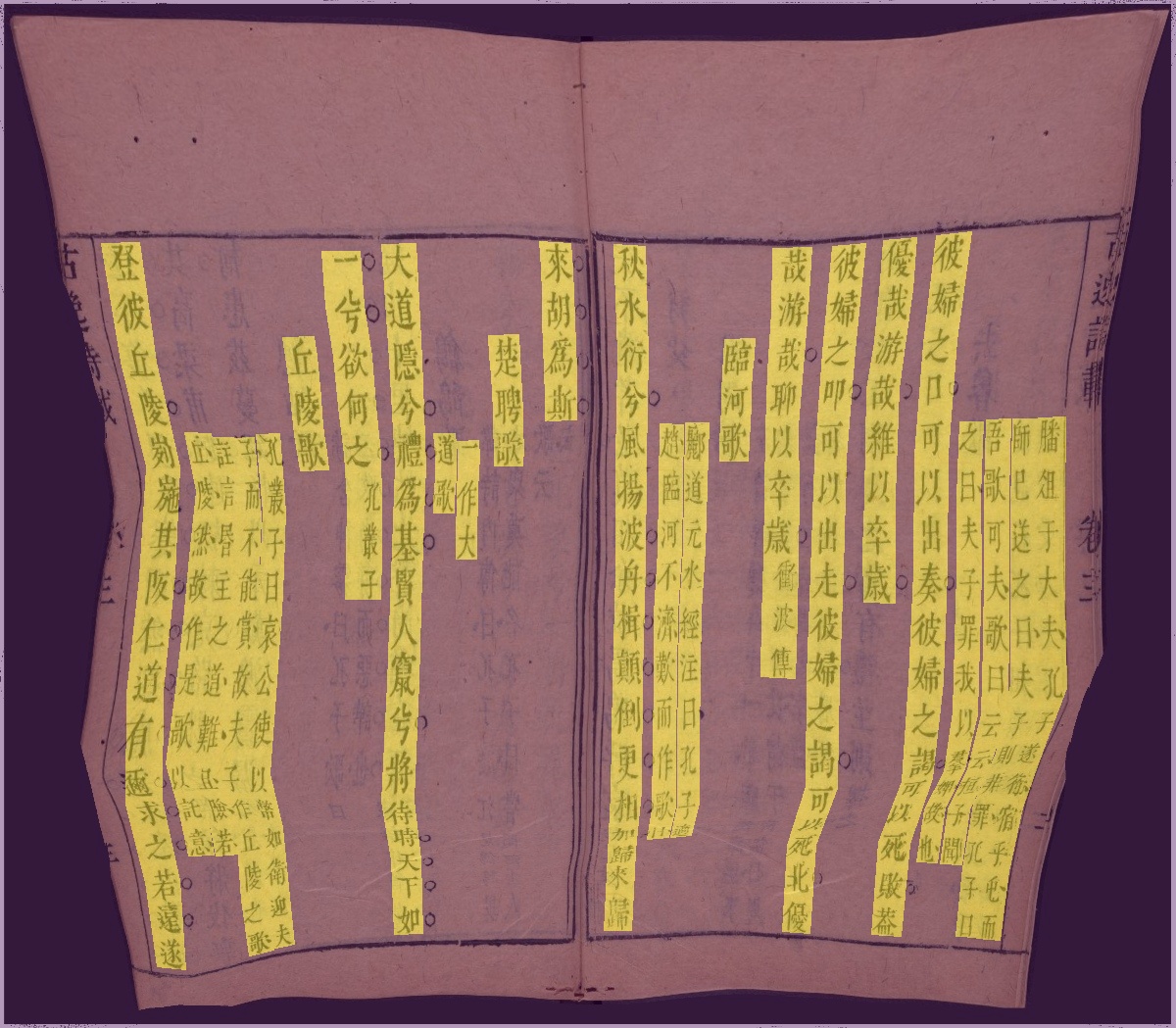}
        \caption{}
        \label{fig:kernel_map_region}
    \end{subfigure}
    \begin{subfigure}{0.24\textwidth}       
        \includegraphics[width=\textwidth]{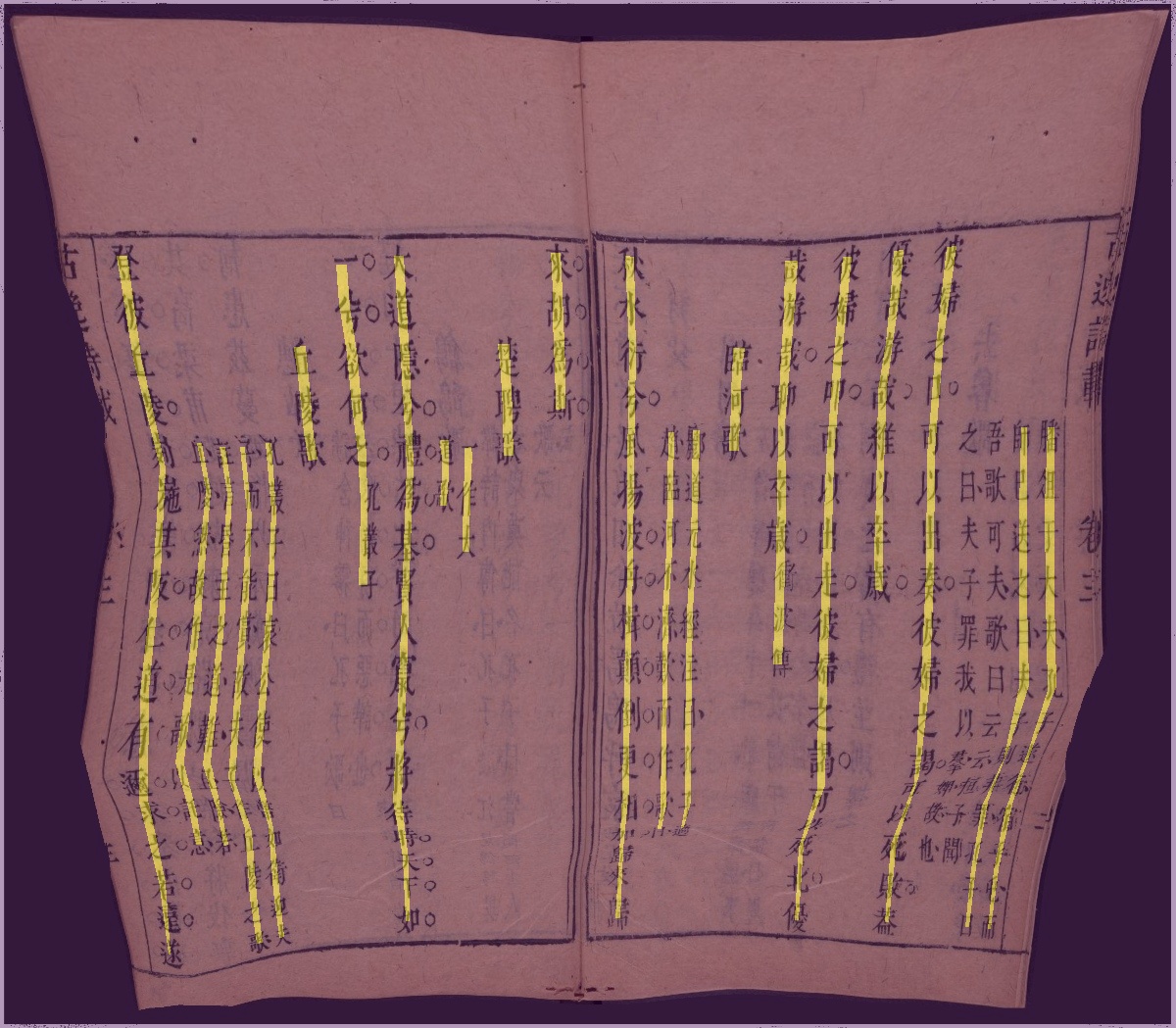}
        \caption{}
        \label{fig:kernel_map_kernel}
    \end{subfigure}
    \hfill
    \begin{subfigure}{0.24\textwidth}
         \includegraphics[width=\textwidth]{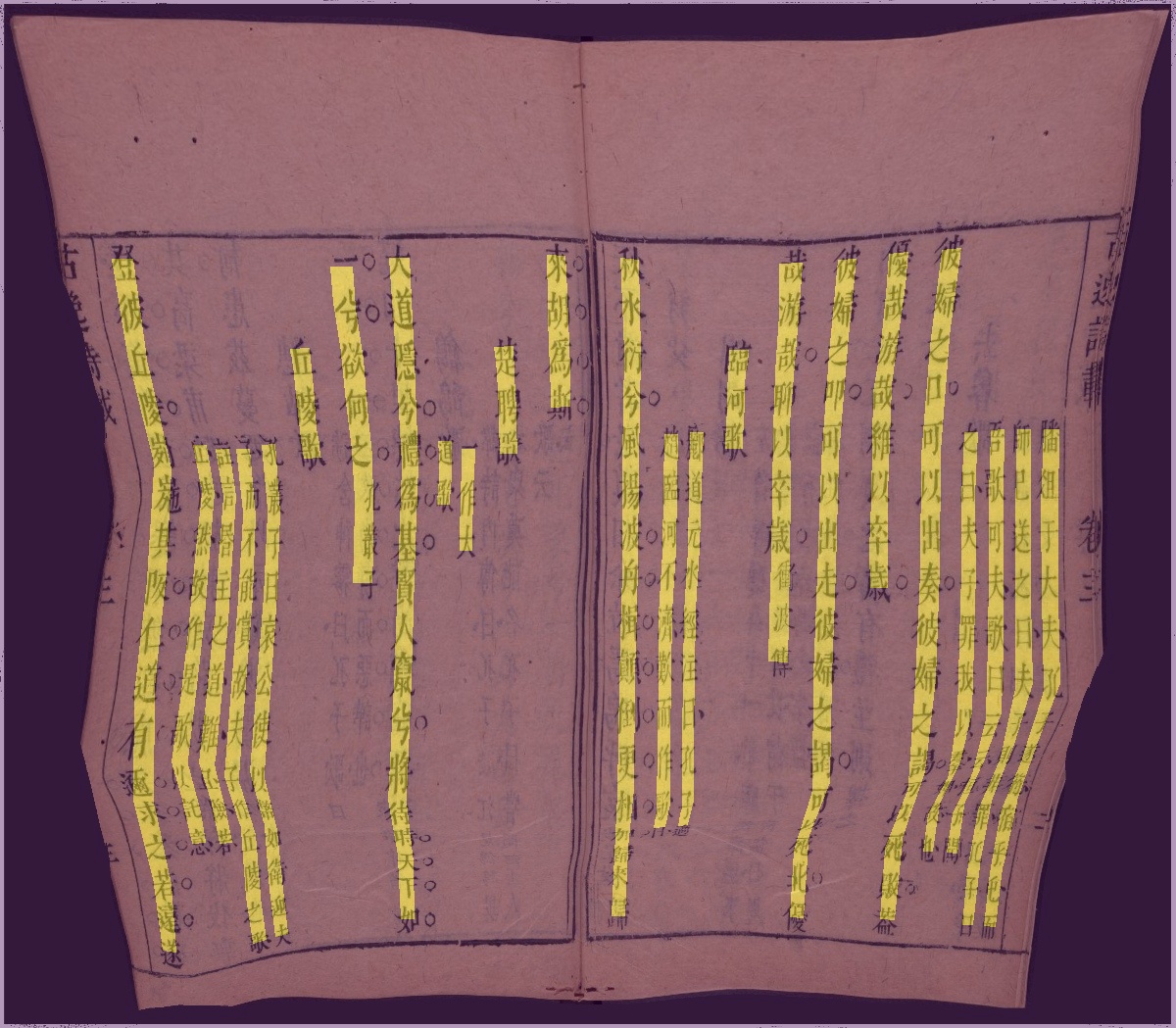}
        \caption{}
        \label{fig:kernel_map_tks}
    \end{subfigure}
    \caption{(a) Historical document from \ac{CHDAC} dataset. (b) Text region map, suffering from overlapping text instances. (c) Text kernel map generated by DB~\cite{liao2020real} with default shrink ratio $r=0.16$, text instances split during shrinkage failed to generate text kernels. (d) Text kernel map generated by \ac{TKS} ($r=0,s=2$).}
    \label{fig:kernel_map}
\end{figure}

To avoid the overlap of text instances, seg-based methods commonly separate nearby text instances by predicting the text kernel map. Text kernel is obtained using Vatti clipping algorithm~\cite{vatti1992generic}, with the shrinkage distance \( D \) calculated using text instance's perimeter \( L \) and its area \( A \) as follows:
\begin{equation}
    D = \frac{A(1-r)}{L},
    \label{eq:tks}
\end{equation}
where \( r \) is the shrink ratio.\footnote{In seg-based methods like \cite{liao2020real}, \( D = \frac{A(1-r^2)}{L} \); here, \( r^2 \) is replaced with \( r \) (where \( r < 1 \)) to achieve a greater shrinking distance.}

\begin{figure}[htbp]
    \centering
    \includegraphics[width=0.7\textwidth]{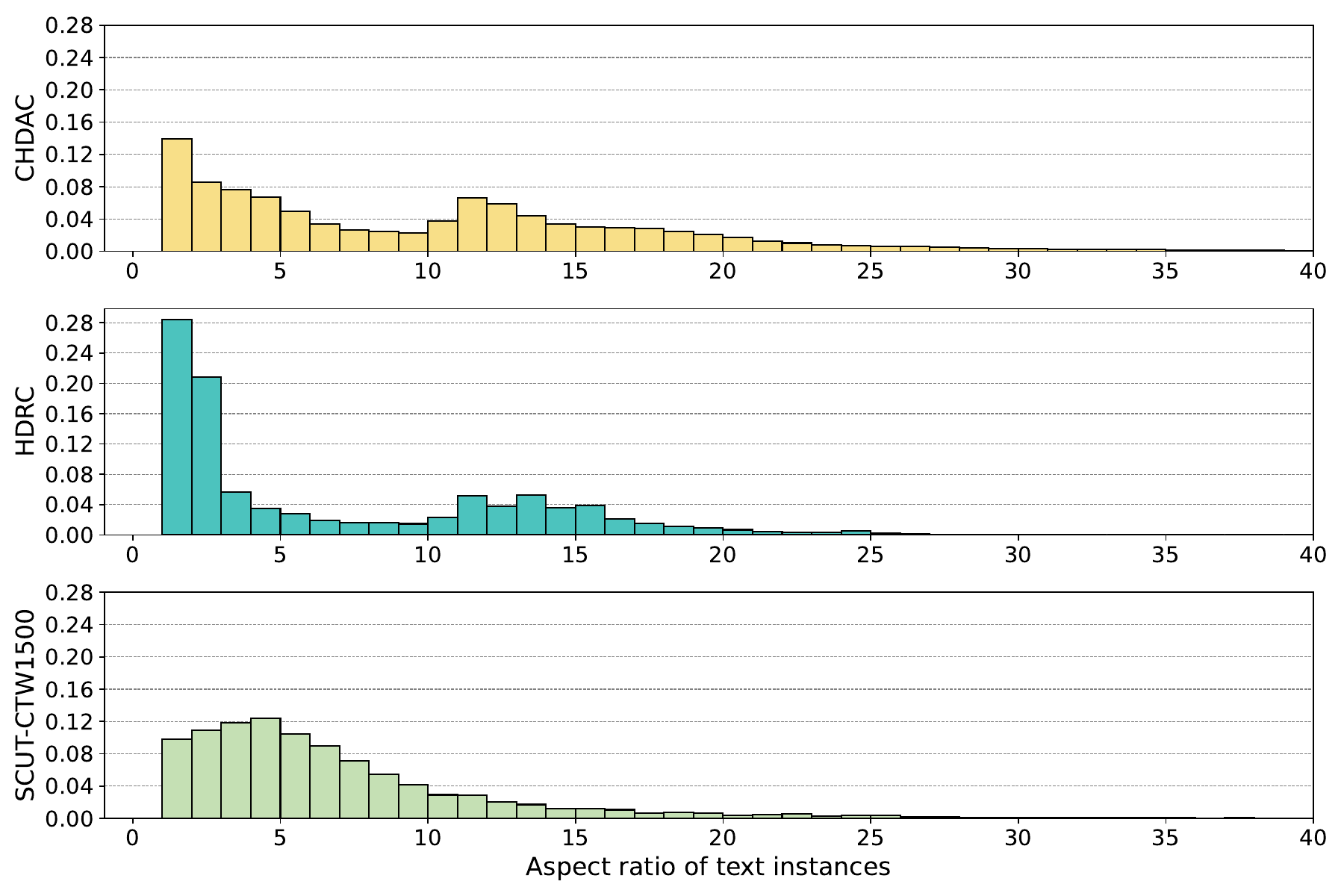}
    \caption{Aspect ratios distribution of text instances in historical document datasets (\ac{CHDAC} and \ac{HDRC}) and in a natural scene dataset (SCUT-CTW1500).}
    \label{fig:distribution}
\end{figure}

The distribution of aspect ratios is depicted in Fig.~\ref{fig:distribution}. Text lines in natural scenes tend to have smaller aspect ratios, whereas historical documents display a bimodal distribution, with some instances having higher aspect ratios. In extreme cases, curved instances with high aspect ratios may split into multiple parts, leading to the failure of text kernel generation, as illustrated in Fig.~\ref{fig:kernel_map_kernel}. Furthermore, text lines in historical document images are densely packed, so separating instances with a large $r$ is impractical, presenting a challenge for target generation.

Considering that the text direction in Chinese historical documents is mainly vertical, for simplicity, \ac{TKS} make the shrinkage along the x-direction only \(1/s\) of that along the y-direction, with \(s \ge 1\) representing the stretch ratio. By adopting different shrinkage distances for the horizontal and vertical directions in this way, it is possible to separate adjacent text lines while avoiding excessive horizontal shrinkage.

\begin{figure}[htbp]
    \centering
    \includegraphics[width=\textwidth]{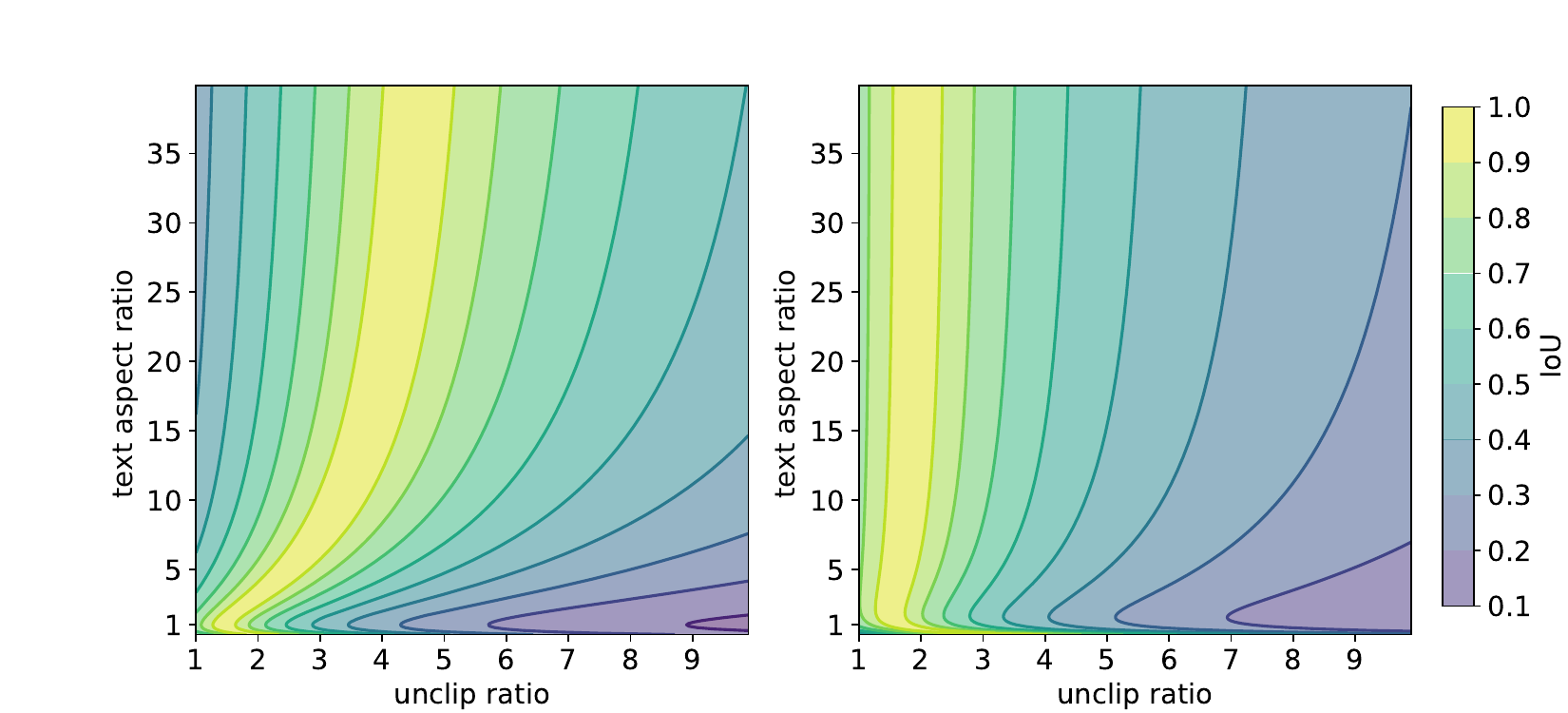}
    \caption{Recovery \ac{IoU} of vertical text kernels with different text aspect ratios using different unclip ratios. Left: text kernels generated by DB ($r=0.16$). Right: text kernels generated by \ac{TKS} ($r=0, s=2$).}
    \label{fig:tks_pp}
\end{figure}

Since seg-based methods predict text kernels, using the post-processing method in DB~\cite{liao2020real} to recover text regions is a simple and direct approach. Our \ac{TKS} targets allow for more accurate recovery by this post-processing method. The method employs the Vatti clipping algorithm~\cite{vatti1992generic} to expand the predicted kernel, and the expansion distance $D'$ is calculated as follows:
\begin{equation}
D' = \frac{A' \times u}{L'},
\label{eq:pp}
\end{equation}
where $A'$ and $L'$ are area and perimeter of kernel, respectively, and $u$ is the unclip ratio. Fig.~\ref{fig:tks_pp} shows the \ac{IoU} obtained by recovering kernels of vertical text lines with different text aspect ratios\footnote{Since \ac{TKS} treats x-direction and y-direction differently, we use \textit{text aspect ratio} instead of the \textit{aspect ratio} to represent the ratio of the length along the text direction (y-direction for vertical text lines) to the text width (x-direction for vertical text lines). The ratio is computed on text lines rather than kernels.} using different unclip ratios. It is hard to find a universal unclip ratio that accurately recovers kernels with different text aspect ratios generated by DB. However, kernels generated by \ac{TKS} can be accurately recovered using unclip ratio \( u=1.5 \).

\subsection{Layout Enhanced Module}

\begin{figure}[htbp]
    \centering
    \includegraphics[width=0.7\textwidth]{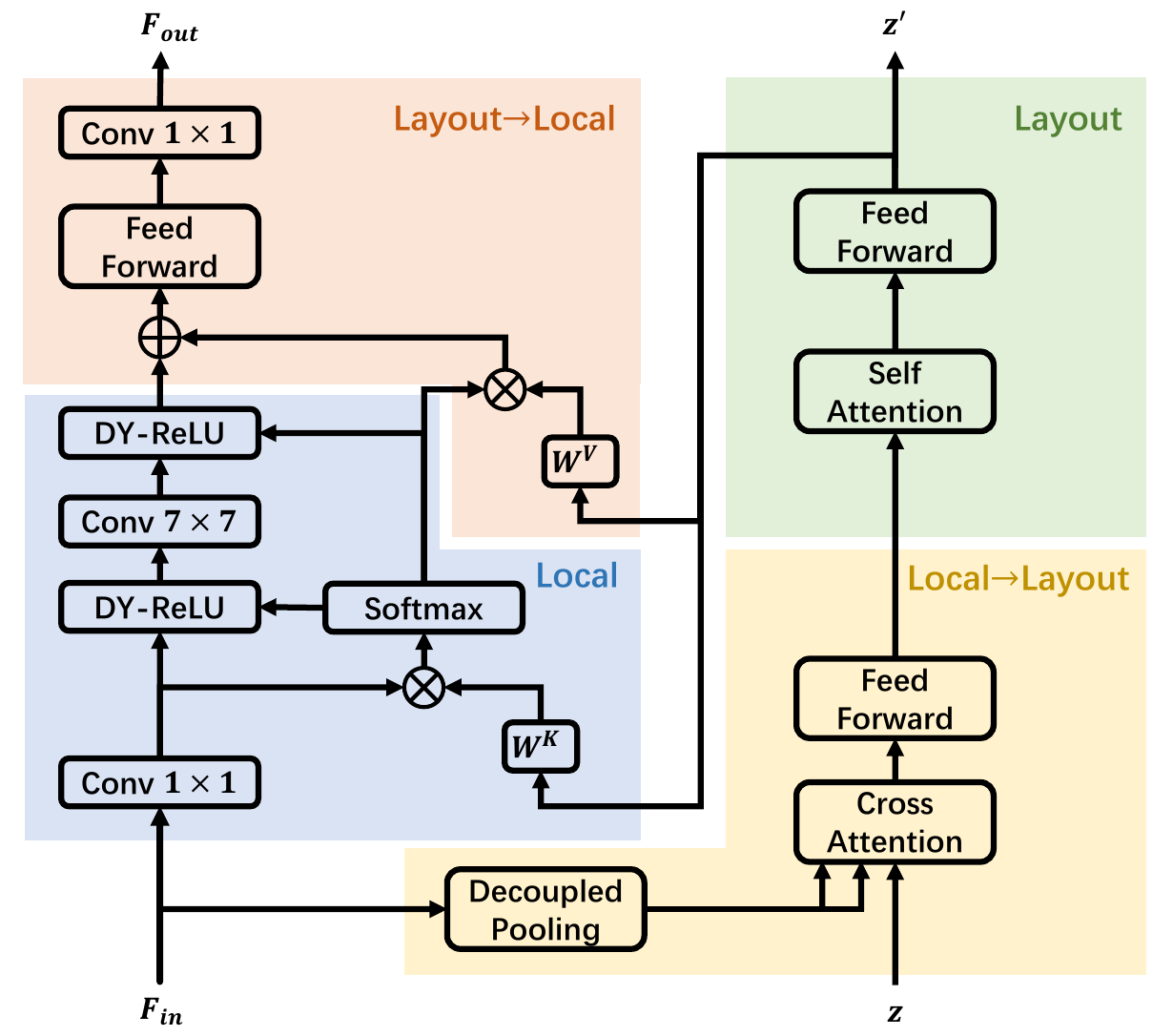}
    \caption{The structure of \ac{LEB}, where the norm module and some residue links are omitted for simplicity.}
    \label{fig:leb}
\end{figure}

To capture the global layout structure without adding excessive parameters and computational load, we design the lightweight \ac{LEM}. Inspired by MobileFormer~\cite{chen2022mobile}, \ac{LEM} consists of multiple \ac{LEB}, whose structure is illustrated in Fig.~\ref{fig:leb}. Each \ac{LEB} takes features \(F_{in} \in \mathbb{R}^{C \times H_0 \times W_0}\) and layout tokens \(z \in \mathbb{R}^{M \times d}\) as input, and outputs the corresponding \(F_{out}\) and \(z'\), each of the same size, respectively. \ac{LEB} comprises four pillars: Local$\to$Layout, Layout sub-block, Local sub-block, and Local$\to$Layout.

\begin{itemize}
    \item \textbf{Local$\to$Layout.}~Local$\to$Layout comprises a multi-head cross-attention and a \ac{FFN}~\cite{vaswani2017attention}. Considering that the layout often appears as rectangular blocks, we perform both horizontal and vertical max pooling on \(F_{in}\), concatenating to form \(\mathbb{R}^{(H_0+W_0) \times C}\), which serves as the keys and values in cross-attention. The layout tokens \(z\) act as the queries. To save computation, we only apply a linear transformation to \(z\) before calculating the attention.
    \item \textbf{Layout sub-block.}~The Layout sub-block adopts the structure of transformer encoder layer~\cite{vaswani2017attention}, resulting in \( z' \in \mathbb R^{M\times d}\).
    \item \textbf{Local sub-block.}~We adopt a bottleneck layer structure~\cite{he2016deep} in Local sub-block. Specifically, we first reduce channel dimension of $F_{in}$, obtaining $F_b \in \mathbb{R}^{C_{b} \times H_0 \times W_0}$. Then, following ConvNext~\cite{liu2022convnet}, we use a $7\times7$ convolution with a group width of 4. After each convolution operation, we apply the spatially-aware DY-ReLU~\cite{chen2020dynamic}, the parameters \( \theta_i \) for pixel \( i \) are obtained as follows:
    \begin{equation}
    \theta_i = \sum_j \alpha_{i,j}f(z'_j), \, \text{where} \, \sum_j \alpha_{i,j} = 1,
    \end{equation}
    where \( \alpha \) represents the attention between \( F_b \) and \( z'\cdot W^K \). To save computation, we use the same \( \theta \) for both DY-ReLU applications.
    \item \textbf{Layout$\to$Local.}~Layout$\to$Local includes multi-head cross-attention and a \ac{FFN}, where attention reuses $\alpha$ computed in the Local sub-block. Lastly, the channels are mapped back to $C$ by a $1\times1$ convolution.
\end{itemize}

Due to the decoupling operation in Local$\to$Layout and the introduction of a bottleneck structure, and by largely avoiding direct attention computation on the deep feature map $F_{in}$, the spatial and temporal computational costs are not significant. We will present the specific statistics on time and number of parameters in Section~\ref{sec:ablation}.

\subsection{Iterative Expansion Distance Post-processor}

\begin{figure}[htbp]
    \centering
    \includegraphics[width=\textwidth]{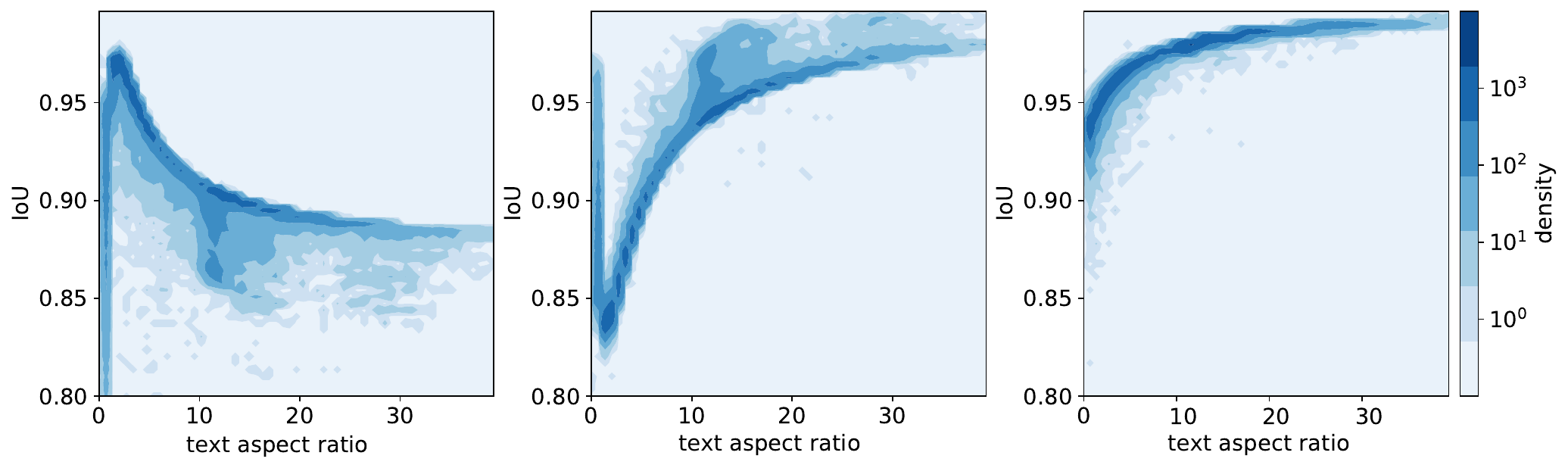}
    \caption{\ac{IoU} recovery levels of text lines in \ac{CHDAC} using different methods, where the text kernels are generated by \ac{TKS} ($r=0, s=2$). Left: DB method ($u=1.5$). Middle: DB method ($u=2.0$). Right: \ac{IEDP}.}
    \label{fig:iedp_recovery}
\end{figure}

As described in Section~\ref{sec:tks}, DB's post-processing method can accurately recover vertical text kernels generated by \ac{TKS}. Illustrated in Fig.~\ref{fig:iedp_recovery}, the recovery of ground truth text kernels in historical documents achieves an \ac{IoU} over 0.8. However, when aiming to build a high-precision detector, such loss of recovery accuracy during the recovery process is undesirable.

Based on the above observation, we propose the \ac{IEDP}. Without significantly increasing the time cost and without introducing hyperparameters, \ac{IEDP} meets the design requirements of high-precision detectors. The procedure is as follows:
\begin{enumerate}[topsep=5pt, partopsep=5pt]
\setlength{\itemsep}{0pt}
\setlength{\parskip}{0pt}
    \item Binarize the predicted text kernel map to obtain the binary map.
    \item Extract connected components from the binary map to serve as text kernels.
    \item For each text kernel, iteratively adjust the expansion distance until the difference between this distance and the newly computed shrinkage distance is sufficiently small. This process is illustrated in Algorithm~\ref{alg:iedp}. Note that the expansion along the x-direction is only $1/s$ of that along the y-direction (i.e. \textbf{expand in \ac{TKS} way}).
\end{enumerate}

\begin{algorithm}
\caption{Iterative Expansion}
\label{alg:iedp}
\begin{algorithmic}[1]
\Procedure{IterativeExpansion}{$kernel, r, s, tolerance$} 
    \State \textbf{Input:} $kernel$ - a text kernel represents by a polygon
    \State \textbf{Input:} $r,s$ - the shrink ratio and the stretch ratio used in \ac{TKS}
    \State \textbf{Input:} $tolerance$ - the tolerance for distance difference
    \State $\textit{d} \gets \Call{InitExpansionDistance}{kernel, r, s}$
    \State $\textit{step}\gets \textit{distance}/2$
    \State $\textit{recovery} \gets \Call{ExpandPoly}{kernel, d, s}$ \Comment{Expand in \ac{TKS} way.}
    \State $\textit{$d'$}\gets \Call{GetShrinkDistance}{recovery, r}$  \Comment{Follow Equation~\eqref{eq:tks}.}
    \While{$|d' - d| > tolerance$}
        \If{$d' >d +t$}
            \State $\text{halve} ~\textit{step}~ \text{if it is greater in last iteration}$
            \State $\textit{d}\gets \textit{d}-\textit{step}$
        \Else
            \State $\text{halve} ~\textit{step}~ \text{if it is less in last iteration}$
            \State $\textit{d}\gets \textit{d}+\textit{step}$
        \EndIf
        \State $\textit{recovery}\gets \Call{ExpandPoly}{kernel, d, s}$
        \State $\textit{$d'$} \gets \Call{GetShrinkDistance}{recovery, r}$
    \EndWhile
\EndProcedure
\end{algorithmic}
\end{algorithm}

By properly setting the tolerance in \ac{IEDP}, the number of iterations can be controlled without significant loss of recovery accuracy, thereby not adding substantial time cost to the post-processing, which will be shown in Section~\ref{sec:ablation}. As shown in Fig.~\ref{fig:iedp_recovery}, using \ac{IEDP}—as opposed to a fixed unclip ratio—can generally achieve a higher level of recovery for text regions with different text aspects ratios, while eliminating the need to adjust the hyperparameters.

\subsection{Integrating Framework to Seg-based Methods}
\label{sec:integrate}

As shown in the ablation study of \cite{liao2020real}, optimizing the text kernel alone often does not yield outstanding results; therefore, SegHist typically needs to be integrated with advanced seg-based methods to achieve better effects. The process of integration with methods (hereinafter referred to as \textit{M}) includes:
\begin{enumerate}
    \item Replace objectives of \textit{M} that involve shrinking or expanding text regions with the objectives generated in \ac{TKS} way, such as the kernel map and threshold map in DB~\cite{liao2020real}.
    \item Insert \ac{LEM} between the feature extractor and prediction head of \textit{M}.
    \item If the post-processing method of \textit{M} only uses a single text kernel map (such as DB~\cite{liao2020real}), use IEDP during inference to recover text regions; otherwise, maintain the original post-processing method (such as progressive scale expansion method of PSENet~\cite{wang2019shape} which uses multiple text kernel maps with different shrinkage distances) to fully utilize the additional prediction targets of \textit{M} to achieve better results.
\end{enumerate}

By integrating the SegHist framework into existing seg-based methods, our approach transfers the performance of advanced seg-based methods from scene text detection to historical document text line detection.

\section{Experiments}

\subsection{Datasets}

\subsubsection{\ac{CHDAC}} The dataset\footnote{Official website of the CHDAC competition: \url{https://iacc.pazhoulab-huangpu.com/}. Similar to HisDoc R-CNN, we only utilized the official dataset and did not use any data submitted by the participants.} comprises a training set of 2000 images and a test set of 1000 images. In this dataset, regular text lines are represented by quadrilaterals, while distorted text lines are annotated using 16 points. Compared to other historical document datasets, text lines are more densely packed and have larger aspect ratios, presenting a greater challenge.

\subsubsection{MTHv2} MTHv2~\cite{ma2020joint} includes \ac{TKH} and \ac{MTH}. The training set contains 2399 images, while the test set consists of 800 images. Its text lines are annotated with quadrilaterals.

\subsubsection{\ac{HDRC}} \ac{HDRC}~\cite{saini2019icdar} is a collection of Chinese family records images. The training set contains 11,715 images and the test set contains 1,135 images. Text lines are annotated using quadrilaterals. Similar to HisDoc R-CNN~\cite{jian2023hisdoc}, we obtained a dataset of 1172 images\footnote{\url{https://tc11.cvc.uab.es/datasets/ICDAR2019HDRC_1}} and randomly divided it into a training set of 587 images, a validation set of 117 images, and a test set of 468 images.

\subsection{Implementation Details}
We integrated the SegHist framework to DB++~\cite{liao2022real}, resulting in DB-SegHist. The DB-SegHist utilized a ResNet-50~\cite{he2016deep} pretrained on ImageNet~\cite{deng2009imagenet} with FPN~\cite{lin2017feature} as backbone. In \ac{TKS}, we used a shrink ratio \(r=0\) and a stretch ratio \(s=2\). Three layers of \ac{LEB} were used to compose the \ac{LEM}, which utilized 8 learnable layout tokens. During training, we used OHEM~\cite{shrivastava2016training} to balance positive and negative samples. For evaluation, we employed the \ac{IEDP} method for post-processing.

Our model was trained on 4 NVIDIA GeForce RTX 2080 Ti GPUs with a batch size of 4, for a total of 30k iterations. We used mixed-precision training~\cite{micikevicius2017mixed}, optimized with AdamW~\cite{loshchilov2017decoupled}, set the initial learning rate to \(1 \times 10^{-4}\), weight decay coefficient to \(1 \times 10^{-4}\), and decayed the learning rate by 0.1 at 50\% and 80\% of the total iterations. Both our training and testing were based on the MMOCR toolkit~\cite{kuang2021mmocr}. 

Additionally, we adopted the data augmentation settings from HisDoc R-CNN for training and testing. Our models were trained using color jittering, flipping, rotating, resizing, and cropping as augmentation techniques. Images were resized to (1333, 800) during testing while keeping their ratio. When assessing the model's robustness to rotated historical document images, we performed random rotations from -15° to 15° on test dataset images from the MTHv2 and the \ac{HDRC}.

\subsection{Ablation Study}
\label{sec:ablation}

\begin{table}[htbp]
    \centering
    \caption{The ablation results of DB-SegHist on the \ac{CHDAC} dataset. Here, $P_{50}$, $R_{50}$, and $F_{50}$ represent the precision, recall, and F-measure, respectively, at an \ac{IoU} threshold of 0.5; $F_{75}$ represents the F-measure at an \ac{IoU} threshold of 0.75. \textit{Time} indicates the average inference time per image on our single GPU setup. All images were resized to $(1333, 800)$ while keeping their ratios.}
    \begin{tabularx}{\textwidth}{*{9}{>{\centering\arraybackslash}X}}
    \toprule 
    Baseline & \ac{TKS} & \ac{LEM} & \ac{IEDP} & \(P_{50}\) & \(R_{50}\) & \(F_{50}\) & \(F_{75}\) & Time (s) \\
    \midrule
    DB++ & & & & 92.43 & 83.89 & 87.95  & 64.91 & 0.0809 \\    
    &  & & $\checkmark$ & 93.68 & 82.57 & 87.77 & 69.05 & 0.0960 \\    
    & $\checkmark$ & & & 98.01 & 94.93 & 96.45 &  90.46 & 0.1082 \\    
    & $\checkmark$ & & $\checkmark$ & 97.98 & 94.90 & 96.42 & 91.67 & 0.1285 \\    
    & $\checkmark$ & $\checkmark$ & & \textbf{98.39} & \textbf{95.90} & \textbf{97.13} & 91.80 & 0.1236 \\    
    & $\checkmark$ & $\checkmark$ & $\checkmark$ & 98.36 & 95.88 & 97.11  & \textbf{92.96} & 0.1394 \\
    \bottomrule
    \end{tabularx}
    \label{tab:ablation}
\end{table}

\subsubsection{Effectiveness of \ac{TKS}} When without \ac{TKS}, we calculated the shrinkage distance using Equation~\eqref{eq:tks}, with $r$ set to 0.16. As shown in Table~\ref{tab:ablation}, the introduction of \ac{TKS} results in significant improvements in various metrics for DB++. Specifically, \ac{TKS} achieves performance gains of 8.50\% and 25.55\% in terms of $F_{50}$ and $F_{75}$, respectively. These results indicate that incorporating \ac{TKS} can effectively adapt to the environment of historical documents.

\subsubsection{Effectiveness of \ac{LEM}} In Table~\ref{tab:ablation}, when comparing \ac{TKS}+\ac{LEM} with \ac{TKS}, under an \ac{IoU} threshold of 0.5, the F-score increases by 0.68\%, with the recall rate increases by 0.97\%. Under an \ac{IoU} threshold of 0.75, the F-score increases by 1.34\%. Furthermore, the total parameters of \ac{LEM} used in the model is 0.95M, with an additional time overhead of approximately 0.013s per image. These confirm its efficiency in terms of both space and time.

\subsubsection{Effectiveness of \ac{IEDP}} Firstly, we tested different unclip ratios on a subset of the training dataset to obtain the average \ac{IoU} when directly recovering text kernels generated by both the standard DB++ and \ac{TKS}. Ultimately, we adopted an $u=2.5$ when using DB++ targets and $u=1.8$ when using \ac{TKS} targets. Please note that when recovering \ac{TKS} kernels with the DB post-processing method, it is required to expand in \ac{TKS} way.

The results in Table~\ref{tab:ablation} indicate that the use of \ac{IEDP} does not lead to a significant improvement when the \ac{IoU} threshold is set to 0.5. However, as the \ac{IoU} threshold increases, under DB++, \ac{TKS}, and \ac{TKS}+\ac{LEM}, the use of \ac{IEDP} results in F-score improvements of 4.14\%, 1.21\%, and 1.16\%, respectively. These results confirm the necessity of \ac{IEDP} for constructing a high-precision text detector. Furthermore, \ac{IEDP} is more conducive to building a universal historical document text detector because it is parameter-free.

\subsubsection{Effectiveness on Different Text Aspect Ratios}
\begin{figure}[htbp]
    \centering
    \includegraphics[width=0.8\textwidth]{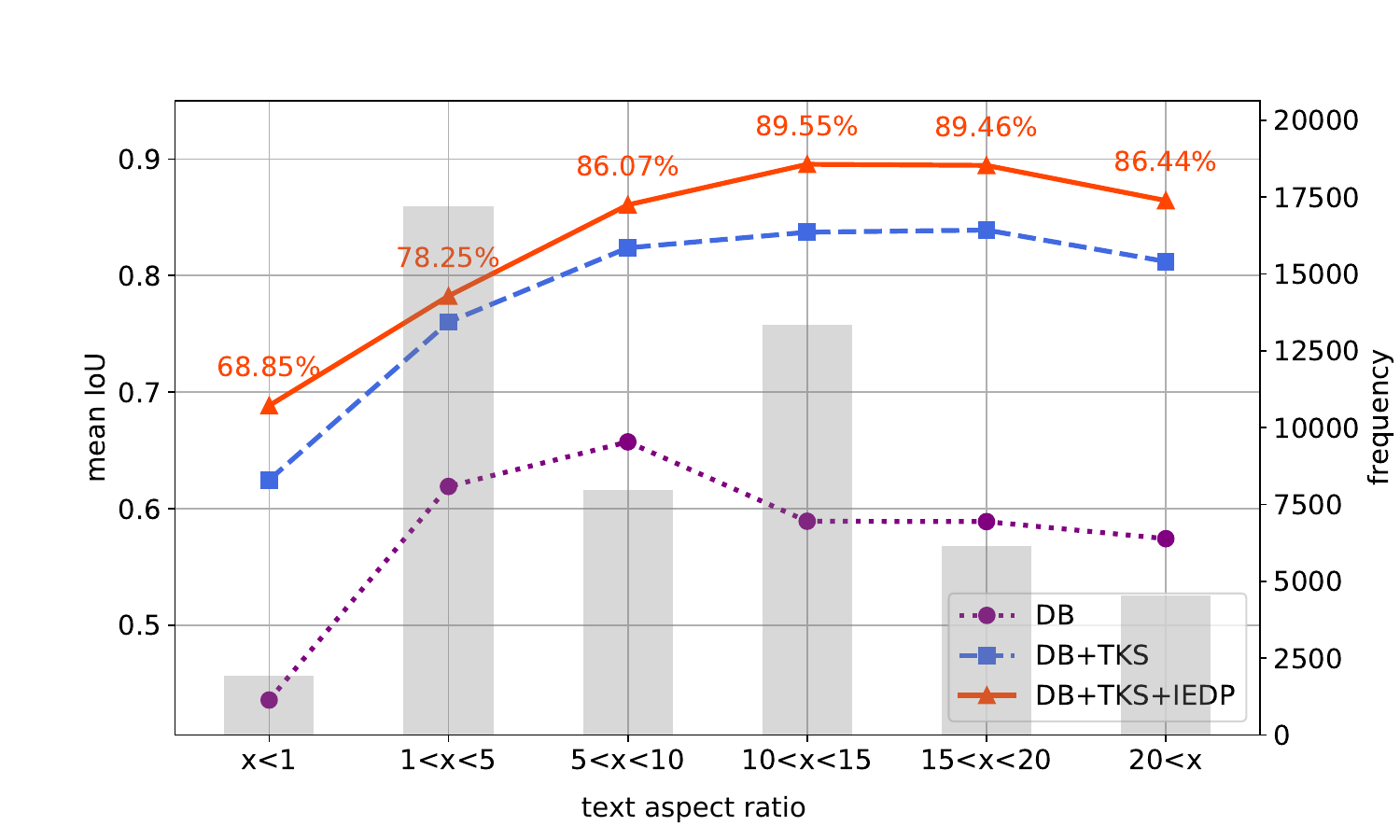}
    \caption{Average \ac{IoU} of text lines with different text aspect ratios on the \ac{CHDAC}. When not using \ac{IEDP} for post-processing, we adhered to DB's default settings with $u=1.5$.}
    \label{fig:aspect_ratio}
\end{figure}

As shown in Fig.~\ref{fig:aspect_ratio}, before introducing \ac{TKS}, the model has low accuracy, and the performance in recovering text lines with high text aspect ratios (e.g. ratios larger than 10) is inferior compared to those with low text aspect ratios (e.g. ratios between 1 and 10). After incorporating \ac{TKS}, the prediction and recovery of text lines become more accurate, with performance on higher text aspect ratios becoming more stable. Subsequently, using \ac{IEDP} for post-processing results in more accurate recovery, especially offering a more significant improvement for text lines with high aspect ratios. These results demonstrate that \ac{TKS} and \ac{IEDP} significantly enhance the capability to detect text lines in historical documents, especially those with high text aspect ratios.

\subsection{Comparison with Previous Methods}

In this section, we evaluate the performance of our approach on all three benchmarks. On the \ac{CHDAC} dataset, we integrated the SegHist framework to PSENet~\cite{wang2019shape} and PAN~\cite{wang2019efficient}, resulting in PSE-SegHist and PAN-SegHist, respectively. As explained in Section~\ref{sec:integrate}, while DB-SegHist used \ac{IEDP} for post-processing, PSE-SegHist and PAN-SegHist retained their original post-processors. Precision, recall, and F-score are all computed at an \ac{IoU} threshold of 0.5.

\begin{table}[htbp]
    \centering
    \caption{Detection results of different methods on \ac{CHDAC}. This dataset includes curved text lines annotated using 16 points. \textit{P}, \textit{R}, and \textit{F} indicate the precision, recall, and F-measure, respectively, at an \ac{IoU} threshold of 0.5. \textbf{Bold} indicates \ac{SOTA} and \underline{underline} indicates the second-best result.}
    \setlength{\tabcolsep}{10pt} 
    \begin{tabular}{cccc}
    \toprule
    Method & P & R & F  \\
    \midrule
    EAST~\cite{zhou2017east} & 61.41 & 73.13 & 66.76\\
    Mask R-CNN~\cite{he2017mask} &  89.03 & 80.90 & 84.77\\
    Cascade R-CNN~\cite{cai2018cascade} & 92.82 & 83.63 & 87.98\\
    OBD~\cite{liu2021exploring} & 94.73 & 81.52 & 87.63\\
    TextSnake~\cite{long2018textsnake} & 96.33 & 89.62 & 92.85\\
    PSENet~\cite{wang2019shape} & 76.99 &89.62& 82.83\\
    PAN~\cite{wang2019efficient}& 92.74& 85.71& 89.09\\
    FCENet~\cite{zhu2021fourier}& 88.42& 85.04& 86.70\\
    DBNet++~\cite{liao2022real}& 91.39& 89.15& 90.26\\
    HisDoc R-CNN~\cite{jian2023hisdoc}& \underline{98.19}& 93.74& 95.92\\
    \midrule
    PSE-SegHist(ours) & 97.00 & \underline{95.31} & \underline{96.15} \\
    PAN-SegHist(ours) & 97.52 & 94.77 & 96.12 \\
    DB-SegHist(ours) & \textbf{98.36} &\textbf{ 95.88} &\textbf{ 97.11} \\
    \bottomrule
    \end{tabular}
    \label{tab:chdac}
\end{table}

\begin{table}[htbp]
    \centering
    \caption{Detection results of different methods on the MTHv2 dataset and the \ac{HDRC} dataset at an \ac{IoU} threshold of 0.5. Text lines in both datasets are annotated by quadrilaterals. * indicates the use of Swin Transformer~\cite{liu2021swin} as backbone, and $^\dagger$ indicates the requirement of character annotations.}
    \setlength{\tabcolsep}{7pt} 
    \begin{tabular}{ccccccc}
    \toprule
    \multirow{2}{*}{Method} & \multicolumn{3}{c}{MTHv2} &  \multicolumn{3}{c}{\ac{HDRC}}\\
    \cmidrule(lr){2-4} 
    \cmidrule(lr){5-7} 
    & P & R & F & P & R & F  \\
    \midrule
    EAST~\cite{zhou2017east} &- &- &95.04 &83.36 &87.70 &85.47\\
    Ma et al.$^\dagger$~\cite{ma2020joint} &– &– &97.72 &– &– &–\\
    Mask R-CNN~\cite{he2017mask} &95.83 &96.35 &96.09 &94.50 &95.11 &94.81\\
    Cascade R-CNN~\cite{cai2018cascade} &\textbf{98.57} &96.52 &97.53 &\underline{94.73} &\underline{95.28} &\underline{95.00}\\
    OBD~\cite{liu2021exploring} &98.17 &97.19 &97.68 &94.45 &94.78 &94.61\\
    TextSnake~\cite{long2018textsnake} &94.31 &91.77 &93.02 &81.70 &72.95 &77.07\\
    PSENet~\cite{wang2019shape} &96.87 &95.82 &96.34 &92.83 &93.68 &93.25\\
    PAN~\cite{wang2019efficient} &97.65 &95.28 &96.45 &93.34 &89.34 &91.30 \\
    FCENet~\cite{zhu2021fourier} &92.47 &88.19 &90.28 &92.38 &91.11 &91.74\\
    DBNet++~\cite{liao2022real} &93.48 &93.22 &93.35 &93.10 &91.05 &92.06\\
    HisDoc R-CNN~\cite{jian2023hisdoc} &\textbf{98.57} &97.05 &97.80 &94.61 &\textbf{95.65} &\textbf{95.13}\\
    Deformable DETR*~\cite{zhu2020deformable}& 97.92& 94.64& 96.25&– &– &–\\
    DTDT*~\cite{li2023dtdt}& 97.94& \textbf{97.86}& \textbf{97.90}&– &– &–\\
    \midrule
    DB-SegHist(ours) & \underline{98.30} & \underline{97.45} & \underline{97.87} & \textbf{94.79} &94.31& 94.55 \\
    \bottomrule
    \end{tabular}
    \label{tab:quad}
\end{table}

\begin{table}[htbp]
    \centering
    \caption{Detection results of different methods on rotated version of MTHv2 and \ac{HDRC} at an \ac{IoU} threshold of 0.5.}
    \setlength{\tabcolsep}{8pt} 
    \begin{tabular}{ccccccc}
    \toprule
    \multirow{2}{*}{Method} & \multicolumn{3}{c}{Rotated MTHv2} &  \multicolumn{3}{c}{Rotated \ac{HDRC}}\\
    \cmidrule(lr){2-4} 
    \cmidrule(lr){5-7} 
    & P & R & F & P & R & F  \\
    \midrule
    EAST~\cite{zhou2017east} &87.01 &89.61 &88.29 &87.46 &89.02 &88.23\\
    Mask R-CNN~\cite{he2017mask}&44.27 &37.65 &40.69 &37.19 &31.25 &33.96\\
    Cascade R-CNN~\cite{cai2018cascade} &60.63 &44.32 &51.21 &42.73 &33.17 &37.35\\
    OBD~\cite{liu2021exploring} &97.49 &84.72 &90.66 &93.94 &88.59 &91.18\\
    TextSnake~\cite{long2018textsnake} &94.46 &88.45 &91.36 &83.96 &69.96 &76.32\\
    PSENet~\cite{wang2019shape} &90.16 &89.70 &89.93 &86.67 &91.49 &89.01\\
    PAN~\cite{wang2019efficient} &97.39 &91.58 &94.40 &92.67 &84.75 &88.53\\
    FCENet~\cite{zhu2021fourier} &89.96 &89.83 &89.89 &89.35 &87.70 &88.52\\
    DBNet++~\cite{liao2022real} &89.92 &90.16 &90.04 &92.95 &90.75 &91.84\\
    HisDoc R-CNN~\cite{jian2023hisdoc} &\underline{98.21} &\underline{96.01} &\underline{97.10} &\underline{94.36} &\textbf{94.35} &\underline{94.36}\\
    \midrule
    DB-SegHist(ours) & \textbf{98.38} & \textbf{96.39}& \textbf{97.38}&\textbf{94.60}& \underline{94.21} & \textbf{94.40}\\
    \bottomrule
    \end{tabular}
    \label{tab:rotate}
\end{table}

\subsubsection{Text Detection on Historical Documents}
The experimental results on the \ac{CHDAC}, MTHv2, and \ac{HDRC} datasets are shown in Tables~\ref{tab:chdac} and Table~\ref{tab:quad}. Our method achieves \ac{SOTA} F-score levels on the CHDAC and MTHv2 datasets under the use of ResNet-50~\cite{he2016deep} as the backbone, surpassing the previous \ac{SOTA} by 1.19\% on the CHDAC dataset. The results of PSE-SegHist and PAN-SegHist on \ac{CHDAC} also show the universality of our framework. Since the text regions in the HDRC dataset are not closely aligned with the actual text areas and contain a considerable number of background pixels, it is challenging for seg-based methods to make accurate per-pixel predictions. The performance of existing seg-based methods on this dataset generally fall short of that achieved by reg-based methods. Compared to other seg-based approaches, our method shows significant improvements by achieving an F-score of 94.55\%, which is comparable to the \ac{SOTA} method's F-score of 95.13\%.

\subsubsection{Text Detection on Rotated Historical Documents}
As shown in the Table~\ref{tab:rotate}, results on Rotated MTHv2 and Rotated \ac{HDRC} demonstrate DB-SegHist's rotational robustness, with F-scores of 97.38\% and 94.40\%, respectively, both reaching \ac{SOTA} levels. In comparison, reg-based methods might require angle prediction to achieve rotational robustness, whereas seg-based methods, due to their pixel-level predictions and text kernel extraction through connected components, inherently perform well in detecting rotated texts. Our SegHist framework inherits this advantage, exhibiting excellent rotational robustness.

\subsection{Limitation}
\begin{figure}[htbp]
    \centering
    \begin{subfigure}{0.35\textwidth}
        \includegraphics[width=\textwidth]{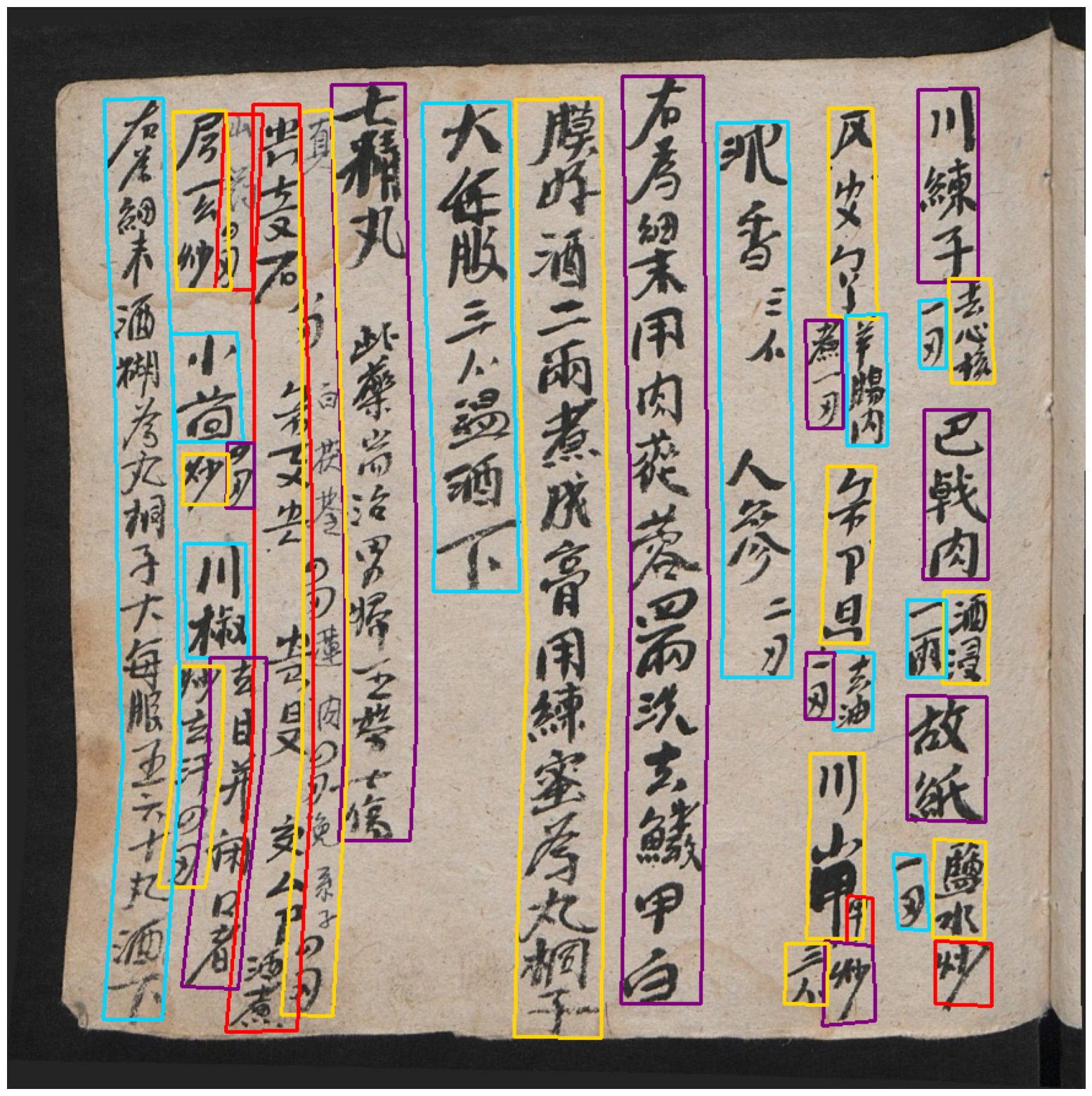}
    \end{subfigure}
    \begin{subfigure}{0.35\textwidth}
        \includegraphics[width=\textwidth]{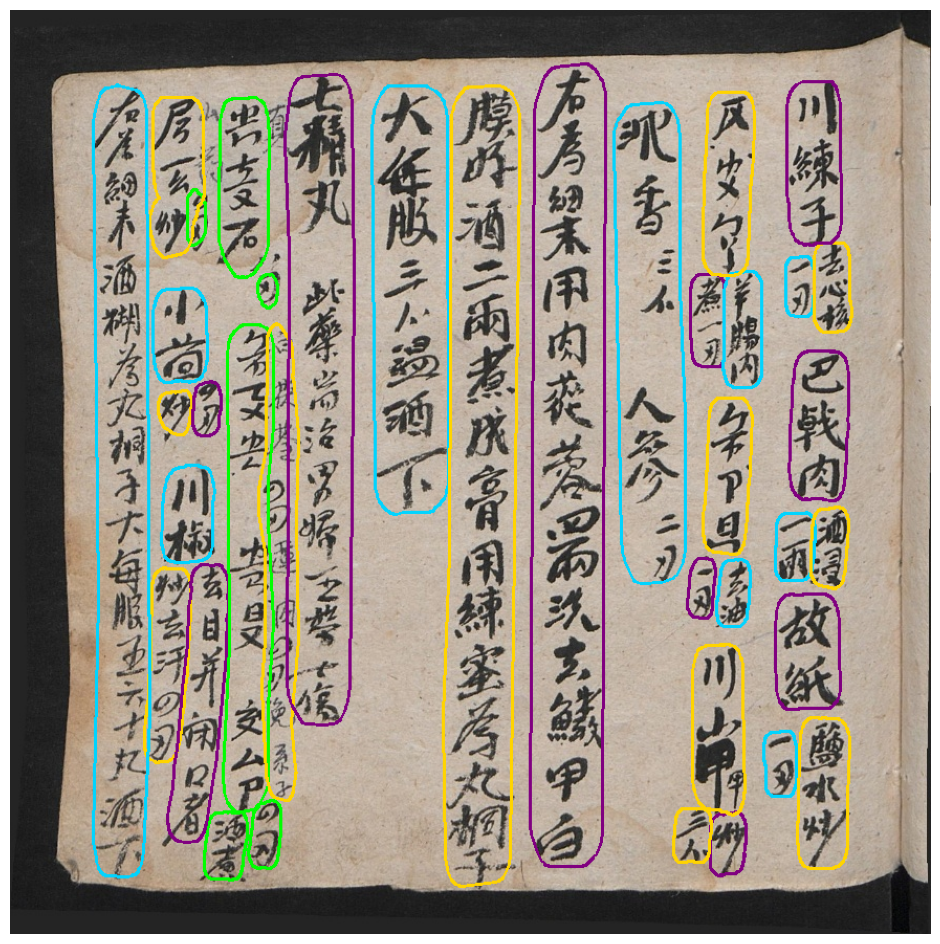}
    \end{subfigure}
    
    \begin{subfigure}{0.35\textwidth}
        \includegraphics[width=\textwidth]{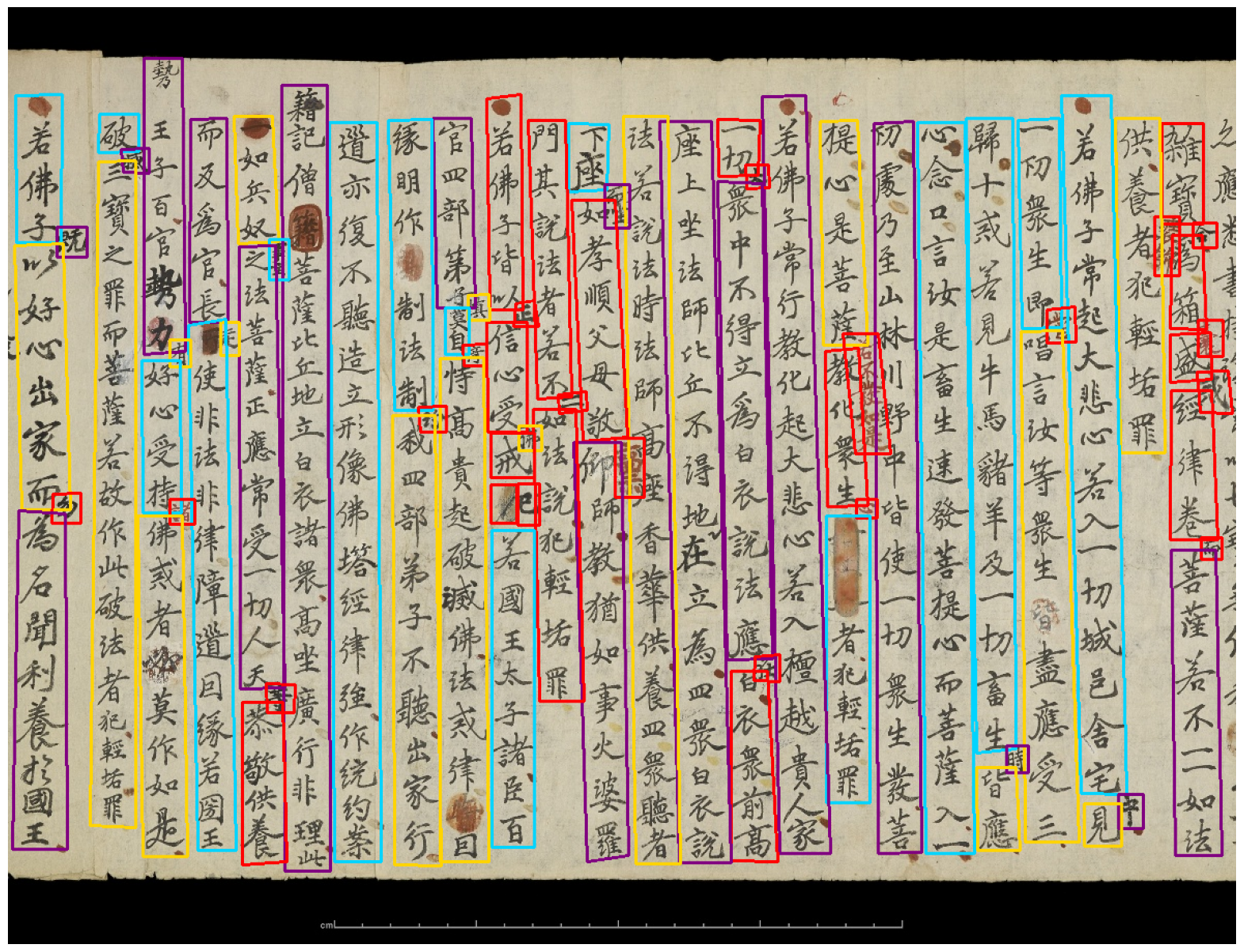}
    \end{subfigure}
    \begin{subfigure}{0.35\textwidth}
        \includegraphics[width=\textwidth]{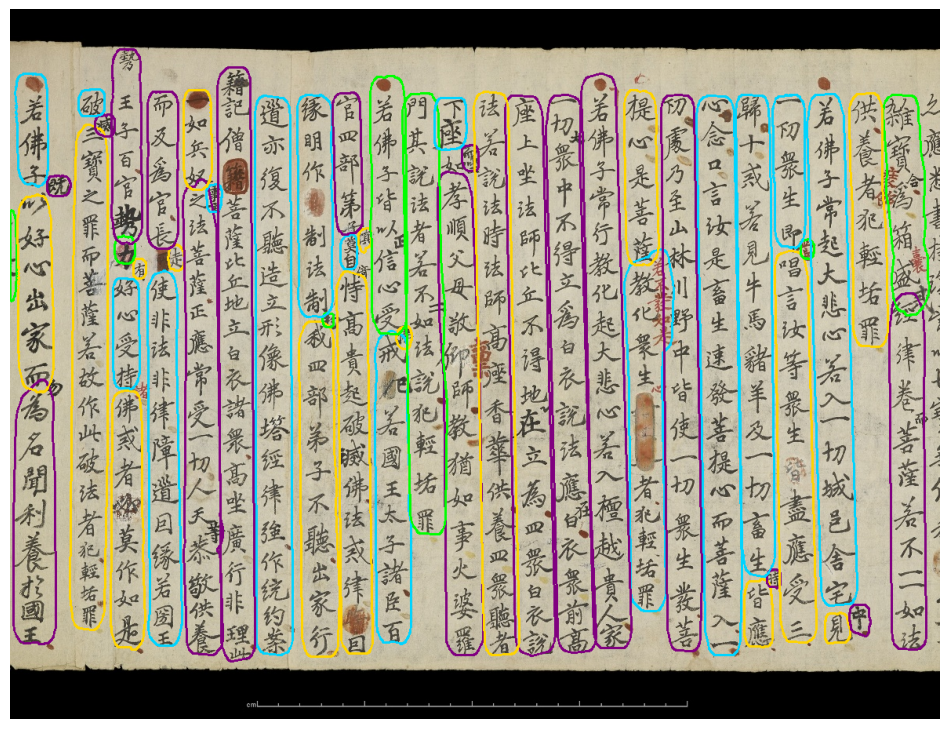}
    \end{subfigure}
    \caption{Examples where our method fails. Left: ground truth. Right: prediction of DB-SegHist. False negatives are marked in red, false positives are shown in green, and other colors are used to display the correspondence between the text lines of the ground truth and the prediction results.}
    \label{fig:case_study}
\end{figure}

Although DB-SegHist demonstrates commendable performance in detecting text in historical documents, particularly for text lines with a text aspect ratio greater than 10, its performance in detecting regions with text aspect ratios less than 5 is still not ideal, as shown in Fig.~\ref{fig:aspect_ratio}. These regions typically appear as annotations or notes, with smaller sizes and fewer characters. As shown in Fig.~\ref{fig:case_study}, although our method is accurate in detecting the main text, it often misses these smaller areas.

Furthermore, since detecting historical documents is closely linked to text recognition, merely considering the F-score of detection is not sufficient. For example, even if a text region is split in prediction, with all characters in the region being detected, it may not result in any loss to the overall \ac{OCR} results, yet it may be considered a complete miss under current metrics. Conversely, for longer text lines, detection may be considered successful even if some characters are missed. The aforementioned phenomenon is not rare and can be observed in Fig.~\ref{fig:case_study}. Thus, the performance of detectors should be evaluated based on their ability to assist in text recognition.

\section{Conclusion}
In this paper, we develop a general framework, SegHist, which extends seg-based methods to historical document data by addressing the challenges of dense texts and text lines with high aspect ratios. Experimental results demonstrate that our DB-SegHist achieves competitive performance across three historical document benchmarks and exhibites strong rotational robustness.

\subsection*{Acknowledgement}
This work is supported by the projects of National Science and Technology Major Project (2021ZD0113301) and the National Natural Science Foundation of China (No. 62376012), which is also a research achievement of the Key Laboratory of Science, Technology and Standard in Press Industry (Key Laboratory of Intelligent Press Media Technology).
%
%
%
\bibliographystyle{splncs04}
\bibliography{ref}
%

\end{document}